# 계층적 어탠션 구조와 트랜스포머를 활용한 알츠하이머 진단과 생성 기반 챗봇

박준영[O], 최창환, 신수종, 이정재[*], 최상일[*]
[O]단국대학교 대학원 인공지능융합학과
단국대학교 SW융합대학 컴퓨터공학과
[*]단국대학교 의과대학 정신건강의학과
[*]단국대학교 SW융합대학 컴퓨터공학과
e-mail: wndwls1024@naver.com[O], ho03206@naver.com, shinsjn@naver.com,
mdjjlee@gmail.com[*], choisi@naver.com[*]

# Alzheimer's Diagnosis and Generation-Based Chatbot Using Hierarchical Attention and Transformer

Park Jun Yeong[*], Shin Su Jong, Choi Chang Hwan, Lee Jung Jae[*], Choi Sang-il[*]
[O]Dept. of AI Convergence, Dankook University
Dept. of Computer Engineering, Dankook University
[*]Department of psychiatry, college of medicine, Dankook University
[*]Dept. of Computer Engineering, Dankook University

### 요 약

본 논문에서는 기존에 두 가지 모델이 필요했던 작업을 하나의 모델로 처리할 수 있는 자연어 처리 아키텍처를 제안한다. 단일 모델로 알츠하이머 환자의 언어패턴과 대화맥락을 분석하고 두 가지 결과인 환자분류와 챗봇의 대답을 도출한다. 일상생활에서 챗봇으로 환자의 언어특징을 파악한다면 의사는 조기진단을 위해 더 정밀한 진단과 치료를 계획할 수 있다. 제안된 모델은 전문가가 필요했던 질문지법을 대체하는 챗봇 개발에 활용된다.

모델이 수행하는 자연어 처리 작업은 두 가지이다. 첫 번째는 환자가 병을 가졌는지 여부를 확률로 표시하는 '자연어 분류'이고 두 번째는 환자의 대답에 대한 챗봇의 다음 '대답을 생성'하는 것이다. 전반부에서는 셀프어탠션 신경망을 통해 환자 발화 특징인 맥락벡터(context vector)를 추출한다. 이 맥락벡터와 챗봇(전문가, 진행자)의 질문을 함께 인코더에 입력해 질문자와 환자 사이 상호작용 특징을 담은 행렬을 얻는다. 벡터화된 행렬은 환자분류를 위한 확률값이 된다. 행렬을 챗봇(진행자)의 다음 대답과 함께 디코더에 입력해 다음 발화를 생성한다.

이 구조를 DementiaBank의 쿠키도둑묘사 말뭉치로 학습한 결과 인코더와 디코더의 손실함수 값이 유의미하게 줄어들며 수렴하는 양상을 확인할 수 있었다. 이는 알츠하이머병 환자의 발화 언어패턴을 포착하는 것이 향후 해당 병의 조기진단과 종단연구에 기여할 수 있음을 보여준다.

▶ Keyword : 알츠하이머(Alzheimer); 생성기반챗봇(Generative chatbot); 자연어 분류(Natural Language Classification)

## I. Introduction

알츠하이머 병은 치매 증상을 가지는 질환 중 60% 이상을 차지하는 대표적 치매 정신 질환이다. 혈관성 치매와 달리 서서히 병의 악화가 진행된다.

환자들에게서 발생하는 특이적 언어 패턴을 감지하면 자기공명영상 (Magnetic Resonance Imaging, MRI)과 함께 병의 조기발견과 치료에 기여할 수 있다. 병이 진행될수록 환자의 언어적 변화가 두드러진다. 인지기능의 저하로 인해 초기에는 기억력 손상이 나타난다. 이로 인해 눈 앞에 보이는 현상이나 사물을 적절한 단어로 설명하지 못하고 경험했던 이야기를 떠올리지 못한다. 중증도가 심해짐에 따라 판단력

이 흐려지고 우울증이나 망상을 동반하며 말 수가 급격히 줄어들고 '모른다'라는 대답의 빈도수가 높아진다. 또한 공격적 언행이나 동문서답, 지시어의 남발 등을 관찰할 수 있다.

## II. Preliminaries

### 1. Related works

자연어 처리를 활용해 알츠하이머를 진단하는 초기 연구[1][2][3][4] 머신러닝 방법론을 적용했다. 알츠하이머 병 환자들의 발화에서 어휘적, 문법적 특징들을 추출한 후 SVM(Support Vector Machine)이나 Skip-gram, Naive Bayes 등을 활용해 환자에서 두드러지는 언어패턴을 발견했다.

머신러닝 기반 학습은 상대적으로 적은 데이터를 통해 빠르게 높은 성능의 모델 학습이 가능하다는 장점이 있다. 그러나 특징을 수작업으로 추출해야 하기 때문에 특징 공학자의 전문성이 모델 성능에 영향을 미친다. 또한 언어변화에 유연하게 적응하지 못한다.

머신러닝 방법의 한계를 보완하기 위해 LSTM, CNN과 어텐션 신경망을 합친 딥러닝 기반 언어 모델을 사용한 연구가 이어지고 있다. [5]에서는 CNN과 양방향 GRU를 사용해 특징을 얻고 어텐션 연산하여 언어패턴을 추출한다.

## III. The Proposed Scheme

본 논문에서는 계층적 셀프어텐션을 통해 문맥벡터를 생성하고 인코더[6]에 입력해 알츠하이머 진단 확률을 계산한다. 또한 디코더[6]에 인코더 은닉벡터를 입력해 다음 대답을 생성한다.

토큰들의 임베딩은 파이토치에서 제공하는 사전학습된 파라미터를 사용했다.

학습을 위해 한 환자의 대화를 여러 세션으로 나눈다. INV는 인터뷰 진행자이고 PAR는 참가자를 의미하는 태그이다. 한 샘플은 한 환자의 인터뷰 전체이다. 진행자의 발화로 대화가 시작된다. 진행자를 기준으로 샘플을 세션으로 자른다. 즉, 한 샘플은 [{INV-PAR-PAR-...}, {INV-PAR-...}] 형태가 된다.

전체적인 구조는 [Fig 1]와 같다. 여기서 최대 문장의 수(첫번째 셀프어텐션 신경망 수)는 10개이다.

인코더 전 까지는 피실험자의 발화만 입력된다. 첫번째 셀프어텐션의 입력은 각각의 문장행렬이다. 피실험자의 발화를 문장으로 나눈 후 문장 내 토큰(단어)을 (1, 768) 차원 벡터로 임베딩한다. 따라서 한 문장은 (문장 내 토큰 수, 768)의 2차원 문장행렬이 된다. 한 문장행렬은 셀프어텐션 하나의 입력값이다.

첫번째 계층의 계산이 끝나면 각 신경망은 한 문장 내의 지역 특징을 담은 벡터를 생성한다. 이 벡터들을 연결해 (문장 개수, 768) 차원의 전체 발화행렬을 생성한 후 두 번째 셀프어텐션 계층에 입력한다.

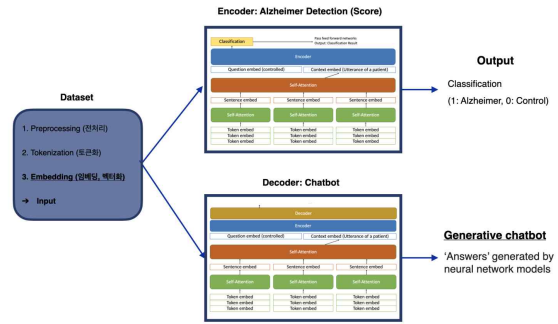

Fig 1. Overview

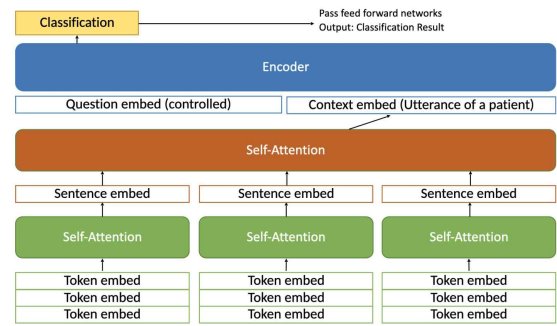

Fig 2. Classification Architecture

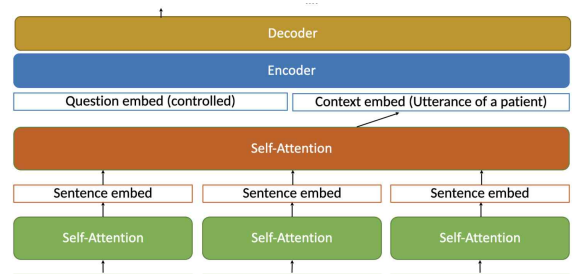

Fig 3. Answer Generation Architecture

두 번째 셀프어텐션 계층을 통해 문장 내외부의 특징을 담은 문맥벡터를 얻을 수 있다. 이 문맥벡터와 진행자의 발화 임베딩을 인코더에 입력한다.

인코더의 결과는 피드포워드 신경망으로 확률이 된다.[Fig 2] 또한 다음 답변과 함께 디코더[Fig 3]에 입력된다. 훈련에서는 디코더의 예측과 다음 세션 진행자의 대답을 비교한다. 추론단계에서는 생성된 발화를 챗봇의 대답으로 출력한다.

## IV. Dataset

본 연구에서는 DementiaBank에서 제공하는 Pit

corpus의 쿠키도둑묘사 영어 대화 말뭉치[7]를 사용했다. 알츠하이머 환자 309명과 대조군 243명을 대상으로 기록되었다. 대화 데이터셋 작성을 위한 .cha 파일을 xml로 변환 후 [Fig 4]의 형태로 정리했다.

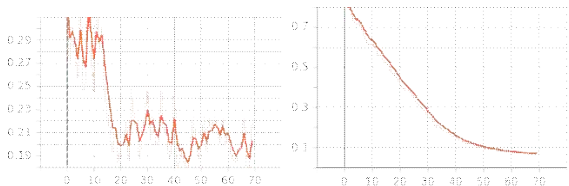

Fig. 4. xml을 파싱하여 csv로 정리

자연어 처리에서는 보통 구두점이나 불용어(특수문자 등)를 제거하지만, 알츠하이머 병 환자의 발화에서는 "umm.."과 같이 불용어로 처리되는 문장들을 특징으로 할 수 있으므로 그대로 남겨둔다.[8]

## V. Conclusions

총 9522개의 문장으로 진행됐으며 배치사이즈 64, 에폭 70으로 설정되었다.

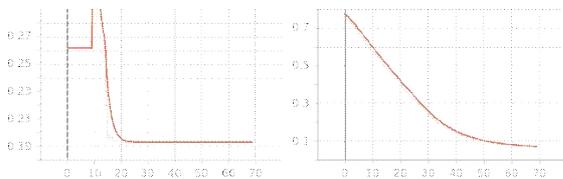

Fig 5. Encoder Loss(left) and Decoder Loss(right) in train

디코더는 비교적 원활하게 수렴되었고 인코더는 20, 50 에폭을 넘어가며 급격히 손실 값이 줄어드는 양상을 확인했다.

Fig 6. Encoder Loss(left) and Decoder Loss(right) in validation

평가 단계에서는 인코더와 디코더 모두 학습이 진행됨에 따라 손실 값이 줄어드는 모습을 볼 수 있다. 평가 단계에서 정확도는 0.738, 에폭 50, 배치 32로 진행시 0.78을 얻을 수 있었다.

본 연구를 통해 계층적인 셀프어탠션 구조로 알츠하이머 병 환자의 언어적 특징을 발견할 수 있다는 점을 보였다. 또한 환자 분류와 대화 생성이 동시에 가능한 End-to-End 구조를 설계할 수 있었다. 추후 연구에서 사전 학습된 모델 가중치를 학습에 사용하고 세션 간 맥락파악을 위한 구조를 적용해 더욱 성능을 향상시킬 수 있을 것이다.

## Acknowledgement

이 논문은 2022년도 정부(과학기술정보통신부)의 재원으로 정보통신기획평가원의 지원을 받아 수행된 연구임(IITP-2022-00155227, 문맥정보를 이용한 딥러닝 기반의 의료 진단에 활용 가능한 ICT-BIO 융합기술 개발, IITP-2017-0-00091, 멀티 모달 딥러닝 기반의 바이오 헬스케어 데이터 분석 기술 개발)